\documentclass[10pt,twocolumn,letterpaper]{article}
\usepackage{bbding}
\usepackage{cvpr}              

\usepackage{graphicx}
\usepackage{amsmath}
\usepackage{amssymb}
\usepackage{booktabs}

\usepackage{multirow}
\usepackage{algorithm}
\usepackage{algorithmicx}
\usepackage{algpseudocode}

\usepackage{xspace}

\usepackage[pagebackref,breaklinks,colorlinks]{hyperref}

\usepackage[capitalize]{cleveref}
\crefname{section}{Sec.}{Secs.}
\Crefname{section}{Section}{Sections}
\Crefname{table}{Table}{Tables}
\crefname{table}{Tab.}{Tabs.}

\makeatletter
\renewcommand{\maketag@@@}[1]{\hbox{\m@th\normalsize\normalfont#1}}%
\makeatother

\def\cvprPaperID{6293} 
\def\confName{CVPR}
\def\confYear{2023}

\begin{document}

\title{Enhanced Multimodal Representation Learning with Cross-modal KD}

\author{Mengxi Chen$^1$, Linyu Xing$^1$, Yu Wang\textsuperscript{1,2\,\Envelope}, Ya Zhang\textsuperscript{1,2\,\Envelope}\\
$^1$Shanghai Jiao Tong University, $^2$Shanghai AI Laboratory\\
{\tt\small \{mxchen\_mc,xly8991,yuwanhsjtu,ya\_zhang\}@sjtu.edu.cn}
}


\maketitle


\begin{abstract}

This paper explores the tasks of leveraging auxiliary modalities which are only available at training to enhance multimodal representation learning through cross-modal Knowledge Distillation (KD). The widely adopted mutual information maximization-based objective leads to a short-cut solution of the weak teacher, \ie, achieving the maximum mutual information by simply making the teacher model as weak as the student model. To prevent such a weak solution, we introduce an additional objective term, \emph{i.e.,} the mutual information between the teacher and the auxiliary modality model. Besides, to narrow down the information gap between the student and teacher, we further propose to minimize the conditional entropy of the teacher given the student. Novel training schemes based on contrastive learning and adversarial learning are designed to optimize the mutual information and the conditional entropy, respectively. Experimental results on three popular multimodal benchmark datasets have shown that the proposed method outperforms a range of state-of-the-art approaches for video recognition, video retrieval and emotion classification.

\end{abstract}

\section{Introduction}
\label{sec:intro}

Multimodal learning has shown much promise in a wide spectrum of applications, such as video understanding~\cite{Shi_2021_ICCV,DBLP:conf/eccv/WangXW0LTG16,DBLP:conf/cvpr/HeilbronEGN15}, sentiment analysis~\cite{DBLP:conf/emnlp/HanCP21,ABDU2021204,DBLP:conf/acl/MorencyCPLZ18} and  medical image segmentation \cite{DBLP:conf/miccai/ChenDJCQH19,4979f9d0d0}. It is generally recognized that including more modalities helps the prediction accuracy.
For many real-world applications, while one has to make predictions based on limited modalities due to efficiency or cost concerns, 
it is usually possible to collect additional modalities at training. 
To enhance the representation learned, several studies have thus considered to leverage such modalities as a form of auxiliary information at training~\cite{dai2021learning,DBLP:conf/acl/ZhaoLJ20,hu2020knowledge,DBLP:conf/aaai/MaRZTWP21}.
Several attempts have explored to fill in the auxiliary modalities at test time, by employing approaches such as Generative Adversarial Network (GAN)~\cite{DBLP:journals/eaai/Ben-CohenKRSBKA19,DBLP:conf/cvpr/HoffmanGD16}. However, such data generation-based approaches introduce extra computation costs at inference. 

\begin{figure}[tp]
    \centering
    \includegraphics[width = 0.46\textwidth]{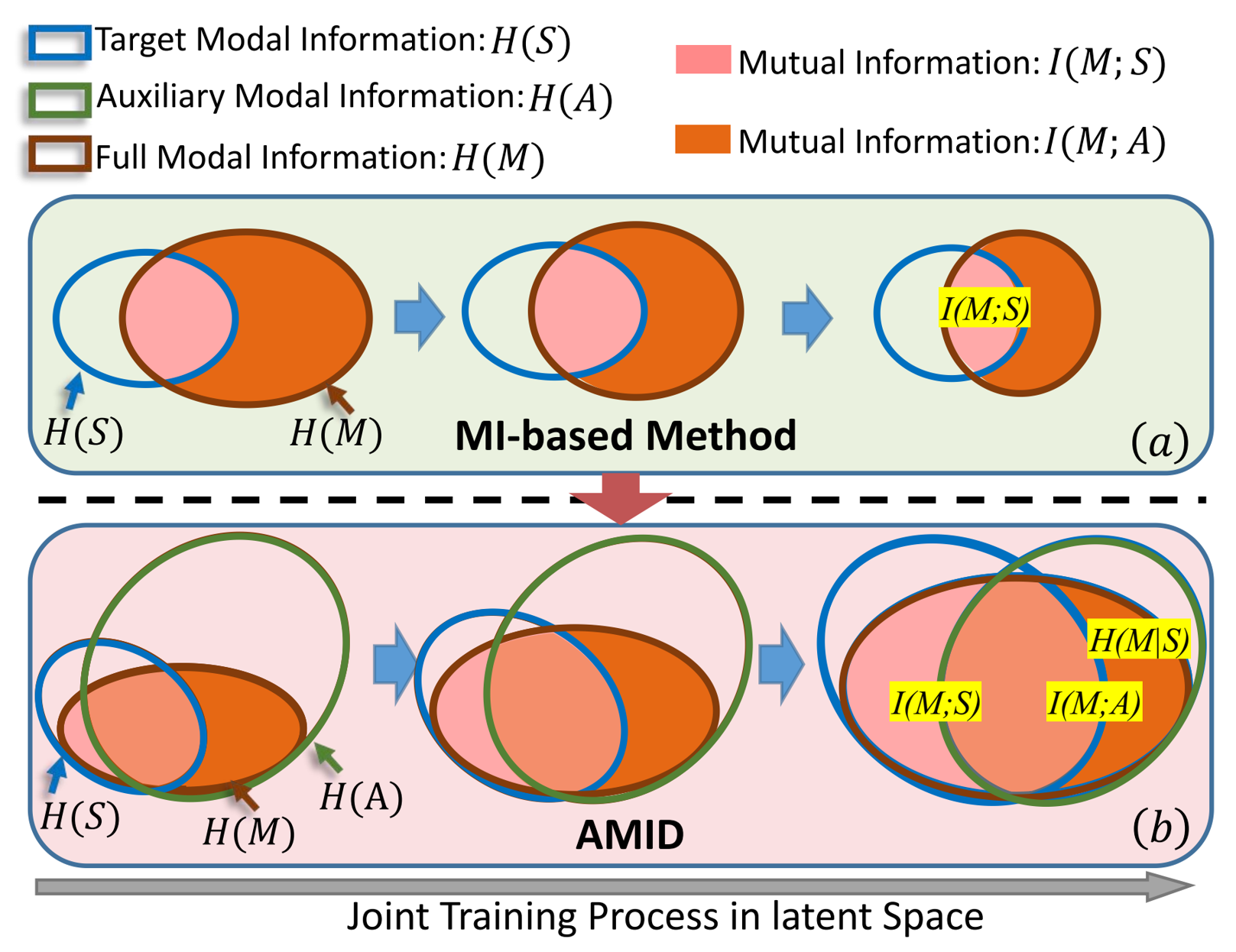}
    \caption{Illustration of the difference between the MI-based method and the proposed AMID method. (a) MI-based methods maximize $I(M;S)$. (b) AMID maximizes $I(M;S)$ and $I(M;A)$, while minimizing $H(M|S)$.
    }
    \label{fig1}
\end{figure}

As an alternative, cross-modal knowledge distillation is proposed, where a teacher model trained with both the target modalities and the auxiliary modalities (denoted as \emph{full modalities} hereafter), is employed to guide the training of a student model with target modalities only ~\cite{DBLP:conf/kdd/XuLGGYPSWSO20,garcia2018modality,xu2020privileged,hu2020knowledge,crasto2019mars,dai2021learning}. In such a way, the student model directly learns to embed information about auxiliary modalities and involves no additional computational cost at inference. A typical practice is to pre-train the teacher model from the full modality data, and fix it during the training. While such a setting enables the teacher to provide information-rich representation~\cite{xu2020privileged,hu2020knowledge}, it may not be favorable for the alignment of the teacher and student, due to the large information gap between them, especially at the early stage of training~\cite{DBLP:conf/aaai/MirzadehFLLMG20}. 


To better facilitate the knowledge transfer, online KD is proposed to jointly train the teacher and the student. One common solution is to maximize the mutual information (MI) between the teacher and the student in a shared space~\cite{DBLP:conf/iclr/TianKI20}.
However, such an objective only focuses on maximizing the shared knowledge, in terms of percentage, from the teacher to the student, which is necessary but not sufficient to guarantee a complete knowledge transfer. As illustrated in~\cref{fig1}(a), a short-cut solution is to simply make the teacher model capture less information, \emph{i.e.}, shrinking the entropy of the teacher. 

Under the framework online KD, this paper explores to simultaneously achieve: 
1) maintaining the representation capacity of the full modality teacher as much as possible, and 2) narrowing down the information gap between the  teacher and the student. As~\cref{fig1}(b) shows, in addition to maximizing the MI between the teacher and the student, maximizing the MI between the teacher and an additional auxiliary model may contribute to the former goal, while minimizing the conditional entropy of the teacher given the student may benefit the latter goal. We thus propose a new objective function consisting of all the above three terms.
To achieve the optimization of the MI, we further derive a new form of its lower bound by taking into account the supervision of the label, and propose a contrastive learning-based approach to implement it. To minimize the conditional entropy, an adversarial learning-based approach is introduced, where additional discriminators are used to distinguish representations between the teacher and the student. We name the proposed method  Adversarial Mutual Information Distillation (AMID) hereafter.


To validate the effectiveness of AMID, extensive experiments on three popular multimodal tasks including video recognition, video retrieval and emotion classification tasks are conducted. The performance of these tasks are conducted on UCF51~\cite{soomro2012dataset}, ActivityNet~\cite{DBLP:conf/cvpr/HeilbronEGN15} and IEMOCAP~\cite{DBLP:journals/lre/BussoBLKMKCLN08} benchmark datasets, respectively. The results show that AMID outperforms a range of state-of-the-art approaches on all three tasks. To summarize, the main contribution of this paper is threefold:
\begin{itemize}
\item We propose a novel cross-modal KD method, Adversarial Mutual Information Distillation (AMID), to prevent the teacher from losing information by maximizing the MI between it and an auxiliary modality model.  
\item AMID maximally transfers information from the teacher to the student by optimizing the MI between them and minimizing the conditional entropy of the teacher given the student.

\item To implement the joint optimization of the three objective terms, a novel training approach that leverages the combined advantage of both contrastive learning and adversarial learning is proposed.

\end{itemize}

\section{Related Work}

\label{sec:related}
{\bf Knowledge Distillation.} 
Knowledge distillation (KD) is first proposed in~\cite{hinton2015distilling}, where soft labels in predictive space produced by the teacher through temperature hyperparameter transfer the dark knowledge to the student. Later, many works studied and demonstrated the benefits of distilling knowledge in the latent space~\cite{DBLP:journals/corr/RomeroBKCGB14,DBLP:conf/iclr/ZagoruykoK17,DBLP:conf/aaai/PassbanWRL21,DBLP:conf/nips/KimPK18}. There have been some works that explore KD-based methods for cross-modal representation learning.  
CMC~\cite{DBLP:conf/eccv/TianKI20} aligns different views of the same samples by contrastive learning to extract shared information. 
CRD~\cite{DBLP:conf/iclr/TianKI20} leverages instance-level contrastive objectives to perform KD.
GMC~\cite{DBLP:conf/icml/PoklukarVYM0K22} learns geometrically aligned representations by aligning modality-specific representations with that of the corresponding complete observation. However, these methods can not transfer the correlation among instances, which are also valuable in many tasks. To transfer the inter-sample relation, \cite{DBLP:conf/cvpr/ChenXKSA21} propose CCL to close the cross-modal semantic gap between the multimodality teacher and the student by contrasting different modal embeddings in the same classes in a common latent space. CRCD\cite{DBLP:conf/cvpr/ZhuTCYLRYW21} distills the information of the teacher space by maximizing two complementary MI criteria. In contrast to these recent KD works which only consider the alignment between the teacher and student but neglect the objectives' impact on the teacher, we proposed a new KD objective that helps the teacher capture as much auxiliary information as possible and constraints the alignment simultaneously.

{\bf Mutual Information Maximization. }MI is first introduced to deep learning in \cite{DBLP:conf/iclr/AlemiFD017}, and since then, many methods have proposed to combine variational bounds with parameterized neural networks to enable differentiable and tractable estimation of MI in high-dimensional spaces \cite{DBLP:conf/nips/FosterJBHTRG19,DBLP:journals/corr/abs-1807-03748}. The character that MI can be used to represent the relationship between two variables is well-suited for correlated multimodal data. \cite{DBLP:conf/emnlp/HanCP21} hierarchically optimizes MI in unimodal pairs and between multimodal and unimodal representations for maintaining useful information while filtering noise in unimodal input. \cite{Liu_2021_ICCV} proposed TupleInfoNCE to optimize MI in multimodal input tuples which enables the model to capture both the modal-shared information and the specific information. However, these methods rarely consider the correlation among samples with the supervision of the label, which are still useful. In this paper, we derive a new lower bound on MI in teacher-student pairs which takes into account the inter-sample correlation and propose a contrastive-learning method for its maximization.

{\bf Adversarial Learning. }Since \cite{DBLP:conf/nips/GoodfellowPMXWOCB14} developed adversarial learning in deep learning, many researchers attempt to improve its performance by stabilizing the training process. In addition to modifying the structure of discriminator or generator~\cite{mirza2014conditional,ChenCDHSSA16,DBLP:journals/corr/RadfordMC15}, several works still considered incorporating multiple discriminators to improve learning \cite{neyshabur2017stabilizing,durugkar2016generative,DBLP:conf/aaai/DoanMAMDPH19}. Similar to \cite{DBLP:conf/aaai/DoanMAMDPH19}, in this paper, we design two additional discriminators and adjust their weights adaptively.


\section{Method}
Suppose we have a multimodal dataset with $n$ modalities,  consisting of $k$ target modalities $\{x_1,\cdots,x_k\}$ and $n$-$k$ auxiliary modalities $\{x_{k+1},\cdots,x_n\}$. Denote the model trained with all $n$ modalities, $k$ target modalities only, and $n$-$k$ auxiliary modalities only as full modality model $M(x_1,\cdots,x_n)$, target modality model $S(x_1,\cdots,x_k)$, and auxiliary modality model $A(x_{k+1},\cdots,x_n)$, respectively.
This paper aims to transfer knowledge from the full modality model $M$ to the target modality model $S$ with the assistant of auxiliary modalities at training. 

A widely adopted objective for cross-modal knowledge transfer is to maximize the MI between the representation of the full modality teacher $M$ and the representation of the target modality student $S$, \ie $I(M;S)$. But there exists a shout-cut solution as shown in~\cref{fig1}(a). In order to prevent this problem, we propose AMID, which maximizes the MI between $M$ and $A$, \ie $I(M;A)$, so as to ensure the full modality model captures information from both target modalities and auxiliary modalities, and maximizes $I(M;S)$ while minimizing the conditional entropy of the teacher given the
student $H(M|S)$, so as to narrow down the information gap between the teacher and the student. For optimizing the MI, we derive a new form of its lower bound and maximize it through a contrastive learning-based approach. For minimizing $H(M|S)$, we introduce an adversarial learning-based approach.

To sum up, the overall objective is to optimize:
\begin{align}
\hspace{-2.2mm}
    \mathcal{O}=\alpha_1 \mathrm{lb}(I(M;S)) + \alpha_2 \mathrm{lb}(I(M;A))- \alpha_3 H(M|S),
\end{align}
where $\mathrm{lb}(\cdot)$ means the lower bound on MI, $\alpha_1$, $\alpha_2$ , $\alpha_3$ are the different weights for the corresponding terms.


\begin{figure*}[tp]
    \centering
    \includegraphics[width=0.95\textwidth]{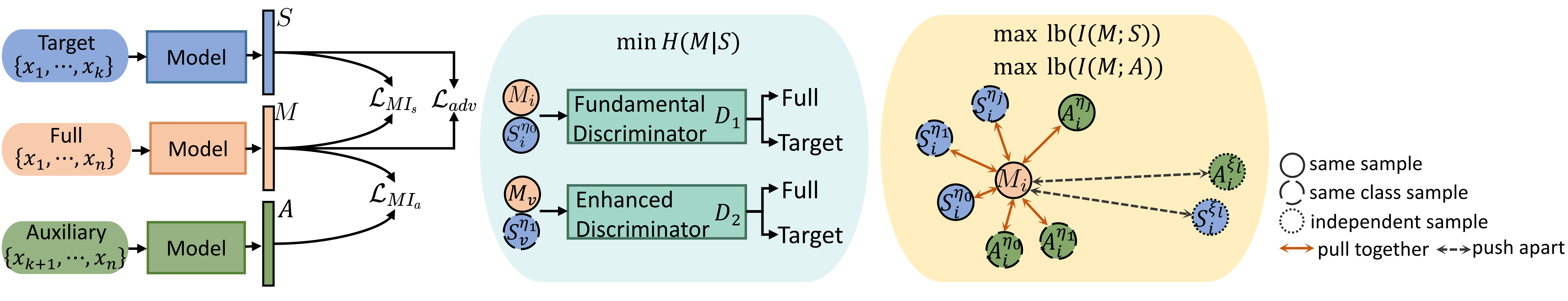}
    \caption{An overview of the proposed AMID. There are three different models, $M$, $S$ and $A$, in AMID for the full modality, target modality and auxiliary modality, respectively.  
    AMID distills multimodal information to the target modality by maximizing the lower bound on $I(M;S)$ and minimizing $H(M|S)$, and prevents the shout-cut solution by maximizing the lower bound on $I(M;A)$. MI optimization is a contrastive-based approach, and conditional entropy minimization is an adversarial learning-based approach that contains two discriminators with different inputs.
    }
    \label{fig2}
\end{figure*}

\subsection{Mutual Information Optimization}
In this section, we will focus on how to optimize $I(M;S)$ and $I(M;A)$ through maximizing their lower bounds for capturing as much information as possible. Unlike~\cite{DBLP:conf/iclr/TianKI20} which only considers the correlation between the teacher's representation and that of the student in the same sample, we derive a new form of lower bound on MI by taking into account the correlation of samples within a class. Notice that as the deduction procedures for $I(M;S)$ and $I(M;A)$ are the same, we will construct an approximate loss for optimizing $I(M;S)$. 

Firstly, we define a distribution $q$ which models the joint distribution of $M$ and $S$. A latent variable, $C$, is used to indicate whether a tuple $(M, S)$ comes from the same class ($C=1$) or not ($C=0$):
\begin{equation}
\begin{split}
    &q(M,S|C=1)= p(M,S); \\ &q(M,S|C=0)= p(M)p(S). 
\end{split}
\end{equation}
We sample $(M,S)$ by considering the relationship between full modality and target modality representations in the same classes, which is more effective for transferring information. The pair $(M,S)$ is from the joint distribution when $M,S$ are in the same class or from the product of the marginal distributions when $M,S$ are in different classes.

Suppose in a mini-batch, given a specific $M_i$, there are $k-1$ positive pairs $(M_i,S^{\eta_j}_i)$ drawn from the joint distribution, where $\{\eta_j\}_{j=1}^{k-1}$ is an index set and every element indicates the position of other $S$ with the same class as $M_i$, $1$ positive pair $(M_i,S^{\eta_0}_i)$ from the same sample, and $N$ negative pairs $(M_i,S^{\xi_l}_i)$ from the product of marginals, where $\{S^{\xi_l}_i\}_{l=1}^N$ are $N$ target modality representations from different classes than $M_i$. Then the priors of $C$ can be calculated as $q(C=1)=\frac{k}{N+k}$ and $q(C=0)=\frac{N}{N+k}$. Using Bayes’ rule, the logarithmic posterior for $C = 1$ is given by:
\begin{align}
    \mathrm{log}q(C=1|M,S)
    &=\mathrm{log}\frac{q(M,S|C=1)q(C=1)}{\sum_{c=0,1}q(M,S|C=c)q(C=c)} \nonumber \\
    &=-\mathrm{log}(1+\frac{N}{k}\frac{p(M)p(S)}{p(M,S)}) \nonumber\\
    &\leq -\mathrm{log}(\frac{N}{k})+\mathrm{log}\frac{p(M,S)}{p(M)p(S)}.
    \label{eq:logq}
\end{align}

With \cref{eq:logq}, we can get a lower bound on $I(M;S)$ as:
\begin{equation}
    \begin{split}
        &I(M;S) = \mathbb{E}_{(M,S) \sim p(M,S)}\mathrm{log}\frac{p(M,S)}{p(M)p(S)} \\
            & \geq \mathrm{log}(\frac{N}{k}) + \mathbb{E}_{q(M,S|C=1)}\mathrm{log}\, q(C=1|M,S).
    \end{split} 
    \label{eq:lowerMI}
\end{equation}
Thus maximizing $\mathbb{E}_{q(M,S|C=1)}\mathrm{log}\, q(C=1|M,S)$ is to maximize the lower bound on $I(M;S)$. However, the true distribution $ q(C=1|M,S)$ cannot be calculated directly. A classification model $h: \{M,S\} \to [0,1]$ is fitted to approximate this distribution. The log-likelihood of the sampled tuples with model $h$ is defined as:
\begin{equation}
    \begin{split}
        \mathcal{I}(h)&=k\mathbb{E}_{q(M,S|C=1)}\left[\mathrm{log}\, h(M,S)\right] \\
        &\quad+N\mathbb{E}_{q(M,S|C=0)}\left[\mathrm{log}\, \left(1-h(M,S)\right)\right].
    \end{split} 
    \label{eq:log-like}
\end{equation}

This likelihood is maximized to make $h$ a good approximation to $q(C=1|M,S)$. Thus, we can get the optimal $h^*(M,S)=q(C=1|M,S)$. Because that both $\mathrm{log}\, h(M,S)$ and $\mathrm{log}\, (1-h(M,S))$ are non-positive for any $h$, we can rewrite \cref{eq:lowerMI} according to \cref{eq:log-like}:
\begin{align}
         I(M;S) &\geq \mathrm{log}(\frac{N}{k}) + \mathbb{E}_{q(M,S|C=1)}\left[\mathrm{log}\, h^*(M,S)\right]\nonumber\\
        &\geq \mathrm{log}\,(\frac{N}{k})+k\mathbb{E}_{q(M,S|C=1)}\left[\mathrm{log}\, h^*(M,S)\right]\nonumber\\
        &\quad +N\mathbb{E}_{q(M,S|C=0)}\left[\mathrm{log}\,(1-h^*(M,S))\right]\nonumber\\
        &=\mathrm{log}\,(\frac{N}{k})+\max_h \mathcal{I}(h)
        \geq \mathrm{log}\,(\frac{N}{k})+ \mathcal{I}(h)
        \label{eq:lower}
\end{align}

Thus we can optimize $I(M;S)$ indirectly by maximizing this lower bound $\mathrm{log}\left(\frac{N}{k}\right)+ \mathcal{I}(h)$. Inspired by~\cite{DBLP:conf/cvpr/ChenXKSA21}, we choose $h$ as:
\begin{align}
\setlength{\abovedisplayskip}{5pt}
\setlength{\belowdisplayskip}{5pt}    h(M_i,S_u)=\frac{\mathrm{exp}\,\left(\Phi\left(M_i,S_u\right)/\tau\right)}{\sum_{j=1}^{N+k} \mathrm{exp}\,(\Phi(M_i,S_j)/\tau)},
    \label{eq:metric}
\end{align}
where $\Phi$ is a cosine similarity function, $\tau$ is a temperature adjusting the correlation of different pairs. Substitute~\cref{eq:metric} into~\cref{eq:lower}, optimizing $I(M;S)$ is equivalent to minimizing the following loss:
\begin{equation}
    \begin{split}
         \mathcal{L}_{MI_s} &= -\frac{1}{N+k}\sum_{i=1}^{N+k}\left( \sum_{j=0}^{k-1}\mathrm{log}\,h(M_i,S^{\eta_j}_i)  \right.\\ &\left. \quad + \sum_{l=1}^N \mathrm{log}\,\left(1-h(M_i,S^{\xi_l}_i)\right) \right).
    \end{split}
    \label{eq:loss MIs}
\end{equation}
Similarly, we can optimize the $I(M;A)$ by minimizing the following loss: 
\begin{align}
    \begin{split}
         \mathcal{L}_{MI_{a}} &= -\frac{1}{N+k}\sum_{i=1}^{N+k}\left(\sum_{j=0}^{k-1}\mathrm{log} \, h(M_i,A^{\eta_j}_i) \right. \\ &\left. \quad +\sum_{l=1}^N \mathrm{log}\,\left(1-h(M_i,A^{\xi_l}_i)\right)\right).
    \end{split}
    \label{eq:loss MIa}
\end{align}

The form of~\cref{eq:loss MIs} and~\cref{eq:loss MIa} are similar to the CCL loss in~\cite{DBLP:conf/cvpr/ChenXKSA21}. However, our formulation is associated with MI and derives a tighter lower bound on it by taking account into the supervision of the label, which in our experiments is found to be more effective.

\subsection{Conditional Entropy Minimization}
Maximizing the lower bound on $I(M;S)$ is insufficient for narrowing the information gap between the teacher and the student as the optimal lower bound does not mean that MI is maximum. This subsection focuses on further transferring knowledge by minimizing the conditional entropy of the teacher given the student $H(M|S)$. 

The posterior $p(M|S)$ is difficult to calculate, thus we define a new distribution $q(M|S)$ to approximate it. Then $H(M|S)$ can be calculated as in~\cite{ChenCDHSSA16}:
    \begin{flalign}
    \hspace{-1.8mm}
            H(M|S)
            &=-\mathbb{E}_{S\sim p(S)}\left[\mathbb{E}_{M'\sim p(M|S)}[\mathrm{log}\, p(M'|S)] \right]\nonumber\\
            &=-\mathbb{E}_{S\sim p(S)}[\mathbb{E}_{M'\sim p(M|S)} \left[\mathrm{log}\,q(M'|S)\right] \nonumber \\ & \quad -D_{KL}\left(p(M|S)||q(M|S)\right) ] \nonumber\\
            &\leq - \mathbb{E}_{S\sim p(S)}\left[ \mathbb{E}_{M'\sim p(M|S)} [\mathrm{log}\,q(M'|S)]\right].
    \end{flalign}
    
Thus we can minimize $H(M|S)$ by maximizing the negative of its upper bound. We achieve this by introducing an adversarial learning-based approach, and using it to maximize $\mathbb{E}_{S \sim p(S)}\left[ \mathbb{E}_{M' \sim p(M|S)} [\mathrm{log}\,q(M'|S)]\right]$ \wrt the parameters of the student network during the generation and minimize $\mathbb{E}_{S\sim p(S)}[\mathbb{E}_{M' \sim p(M|S)} [ \mathrm{log}\,q(M'|S)] - \mathrm{log}\,p(M|S) ]$ \wrt the parameters of the discriminators. 
In other words, we use the discriminator $D(\cdot)$ to recognize whether features come from multimodal distribution or not and feedback informative gradient to the student model through minimizing:
\begin{equation}
\setlength{\abovedisplayskip}{5pt}
\setlength{\belowdisplayskip}{5pt}
    \mathcal{L}_{adv}=-\mathbb{E}_{S \sim p(S)}\left[\mathbb{E}_{M'\sim p(M|S)} [\mathrm{log}\,D(S)]\right].
    \label{eq:loss adv}
\end{equation}
Minimizing \cref{eq:loss adv} \wrt the parameters of the target modality student enables the student to imitate representations coming from the multimodal distribution. It is worth noting that optimizing $H(M|S)$ through~\cref{eq:loss adv} does not affect the learning of the teacher since no gradient back propagates to the teacher.

Nevertheless, for the samples of the same class, it is more challenging to transfer the multimodal information by adversarial learning. This is because the generated target modality representation tends to mimic that of multimodality in the same instance with the constraint of MI, easier but can not capture the correlation among samples with the supervision of the label. 
As a result, we introduce two discriminators to learn more informative target modal representations. One discriminator called fundamental discriminator ($D_1$) is designed to distinguish the target modality and full modality representations from the entire training data. The other discriminator, which is called enhanced discriminator ($D_2$), is designed to further distinguish the two kinds of representations coming from the classes with more than one sample in a batch since we think paying attention to these representations can help the student further aggregate representations in the same class.
Thus, minimizing \cref{eq:loss adv} \wrt the parameters of the student becomes minimizing:
\begin{small}
\begin{flalign}
\mathcal{L}_{adv}\!=\!-\frac{\lambda_1}{N+k}\!\sum_{i=1}^{N+k}\!\mathrm{log}\,D_1(S^{\eta_0}_i)
        \! - \! \frac{\lambda_2}{d}\!\sum_{v=1}^{w}\!\mathrm{log}\,D_2(S^{\eta_1}_v).
\end{flalign}
\end{small}
And the parameters of discriminators are trained by minimizing: 
\begin{small}
\begin{equation}
    \begin{split}
        \mathcal{L}_{D}\!=\!&\!-\!\frac{\lambda_1}{N+k}\!\sum_{i=1}^{N+k}\!\left(\mathrm{log}\,D_1(M_i)\!+\!\mathrm{log}\,\left(1\!-\!D_1(S^{\eta_0}_i)\right)\right)\\ 
        &\!-\! \frac{\lambda_2}{d}\!\sum_{v=1}^{w}\!\left(\mathrm{log}\,D_2(M_v)\!+\!\mathrm{log}\,\left(1\!-\!D_2(S^{\eta_1}_v)\right)\right),
        \label{eq: discriminator}
    \end{split}
\end{equation}
\end{small}
where $\{M_v\}_{v=1}^w$ is a set where every element comes from the classes with more than one samples and $\{S^{\eta_1}_v\}_{v=1}^w$ is its corresponding set of target modal representations. 
$d$ is used for more stabilized training and we set it by the frequency of the data with samples of the same classes. $\lambda_1$ and $\lambda_2$ are weighting parameters for $D_1$ and $D_2$ respectively, and $\lambda_1+\lambda_2=1$. To further improve the efficiency of the transfer, we adjust $\lambda_1$ and $\lambda_2$ adaptively according to the alignment between $S$ and $M$ for each epoch $t$. 
More specifically, we use the averaged cosine similarity in the tuples $\Phi(M_i,S_i^{\eta_0})$ and $\Phi(M_v,S_v^{\eta_j}), j\neq0$ in \cref{eq:metric} to provide meaningful feedback for the cooperation of discriminators. 
Large cosine similarity indicates that $M$ and $S$ are well aligned, thus the weight of the corresponding discriminator can be relatively small. Moving average \cite{Matiisen2020Teacher} is also applied to improve the training stability. The formula of $\lambda$ is given as:
\begin{equation}
\hspace{-2mm}
    \begin{aligned}
    \tilde{\lambda}(t+1) &= \beta\tilde{\lambda}(t) + \left(1-\beta \right)\left(1-\mathrm{Avg}_{t}\left(\Phi(M,S)\right)\right)) \\
    \lambda(t+1) &= \sigma(\tilde{\lambda}(t+1)),
    \label{eq:lambda}
    \end{aligned}
\end{equation}
where $\sigma(\cdot)$ is the softmax function, $\mathrm{Avg}_{t}(\cdot)$ represents the average of cosine similarity on epoch $t$, and $\beta \in (0,1)$ is the smoothing parameter, and we set $\beta = 0.9$ empirically. 

Adversarial learning is known to often generate random images if not constrained~\cite{mirza2014conditional}. This also occurs when target modality representations are generated without any constraints. Our method implicitly overcomes this problem by maximizing the lower bound on $I(M;S)$ and its advantages come twofold: First, the maximization of the lower bound on MI can not only transfer the multimodal information but also place a constraint on the adversarial distillation. Second, adversarial distillation further reduces the information gap that still exists after maximizing the lower bound on MI. Thus, the maximization of the lower bound on MI and adversarial distillation are complementary to each other in AMID to generate informative and robust representations. 

\subsection{Overall Loss}
Assuming that after the optimization above, the target modality and full modality representations are already aligned, then we can use two classifiers with shared parameters to perform the classification task for these representations respectively. The overall classification loss is:
\begin{equation}
    \mathcal{L}_{cls}=\mathcal{L}_{ce}^S(S_i,y_i) + \mathcal{L}_{ce}^M(M_i,y_i),
    \label{eq:loss cls}
\end{equation}
where $\mathcal{L}_{ce}$ is the cross-entropy loss, $y_i$ is the ground-truth. Classifiers with the same parameters can also help facilitate the generation of representations from the target modality network to imitate those of multimodality. In addition, the target modality can be aligned with the full modality in the predictive space. We use the Jensen–Shannon divergence (JSD)~\cite{Lin1991Divergence} to optimize the MI between the predictive target modality distribution $(P_S)$ and the multimodality distribution $(P_M)$:
\begin{equation}
    \mathcal{L}_{JSD}=\frac{1}{N+k}\sum_{i=1}^{N+k}JSD(P_{S_i}||P_{M_i}).
    \label{eq:JSD}
\end{equation}


The final loss function for distilling knowledge from auxiliary modalities is given as:
\begin{equation}
    \mathcal{L}= \mathcal{L}_{cls} + \mathcal{L}_{JSD} + \alpha_1 \mathcal{L}_{MI_s} + \alpha_2 \mathcal{L}_{MI_a} + \alpha_3 \mathcal{L}_{adv},
    \label{eq:objective}
\end{equation}
The whole procedure is shown in the supplementary and the overall model structure is shown in~\cref{fig2}.

\section{Experiments}
\label{sec:exp}
\subsection{Datasets}
\textbf{Video understanding.}
We experiment on two popular multimodal datasets for video recognition (Top-1 accuracy, \%) and kNN video retrieval (R@K, \%)~\cite{DBLP:conf/eccv/BuchlerBO18}. Cross-modal knowledge distillation is performed from the acoustic to the visual modality. 
\begin{itemize}
    \item \textbf{UCF51}, as a subset of UCF101~\cite{soomro2012dataset}, includes 6,845 videos with audio belonging to 51 categories. The public split1 is adopted. 
    \item \textbf{ActivityNet}~\cite{DBLP:conf/cvpr/HeilbronEGN15} is a large-scale video benchmark of 14,950 videos with audio, covering 200 complex human activities, where the training set has 10,024 videos while 4,926 videos in the validation set. 
\end{itemize}

\textbf{Emotion classification.}
We conduct experiments on IEMOCAP~\cite{DBLP:journals/lre/BussoBLKMKCLN08}, a popular multimodal dataset for 4-class emotion classification, which contains totaling 5531 videos in 5 dyadic conversation sessions. Each video consists of acoustic, textual and visual modalities, and the target model is based on visual modality only. A 5-fold cross-validation is adopted for the evaluation, taking one session as the test set for each fold.  
Following~\cite{DBLP:conf/acl/ZhaoLJ20}, two evaluation metrics, weighted accuracy (WA, \%) and unweighted accuracy (UA, \%), are adopted.

\subsection{Experimental Setup}
Because different modalities such as audio, video and text are heterogeneous, we use individual extractors for each modality, then the features are concatenated and passed through a 2-layer MLP as the full modality teacher. 
The teacher shares the same extractors as that of the student and the auxiliary model for reducing the number of parameters.
\textbf{Video recognition and retrieval.} Following \cite{DBLP:conf/cvpr/ChenXKSA21}, we use a 1D-CNN14~\cite{DBLP:journals/taslp/KongCIWWP20} pre-trained on AudioSet~\cite{7952261} as the audio extractor and use R(2+1)D-18 \cite{DBLP:conf/cvpr/TranWTRLP18} as the student video network. The two discriminators are 5-layer MLP and 3-layer MLP respectively. Besides, we change all activation functions to Leaky~ReLU for adversarial learning because it gives better training stability compared to the ReLU. The dimension of the representations is set to 512. 
The batch size and temperature $\tau$ are set to $16$ and $0.5$, respectively. $\alpha_1,\alpha_2,\alpha_3$ are set to $4,1.6,1$.
More details are shown in the supplementary. 

\textbf{Emotion classification.} We extract raw acoustic, textual and visual embeddings by the pre-trained wav2vec~2.0~\cite{DBLP:journals/corr/abs-2006-11477}, BERT~\cite{DBLP:conf/naacl/DevlinCLT19} and DenseNet~\cite{DBLP:conf/cvpr/HuangLMW17} respectively. RNN is adopted as the student network with a final feature dimension of $768$. The batch size and temperature $\tau$ are set to $128$ and $0.6$, respectively. 

A warm-up stage is employed for all experiments. At the first $T_{start}$ epochs, the student model and the teacher model are trained by minimizing $\mathcal{L}=\mathcal{L}_{cls} + \mathcal{L}_{JSD}$ to alleviate the converge difficulties of adversarial learning. This helps a preliminary alignment between the teacher and student and can also prevent discriminators from recognizing readily at the beginning of the training. $T_{start}$ is set to 20 for the video understanding tasks and 5 for the emotion classification task. And notice that without special mention, the leveraged 
pre-trained auxiliary modality extractors are fixed since the generated representations already contain rich auxiliary information.

\begin{table}[t]
\setlength{\abovecaptionskip}{0cm}
\centering
\caption{The Top-1 accuracy (\%) of video recognition on UCF51 and ActivityNet. Notice that we change all the activation functions to Leaky ReLU to ensure a fair comparison.}
\setlength{\tabcolsep}{1.2mm}
\scalebox{1.0}{
\begin{tabular*}{0.7\linewidth}{l|c|c}
    \hline
                                         & UCF51         & ActivityNet   \\ \hline
Baseline                                      & 66.7          & 48.8          \\ \hline
CRD~\cite{DBLP:conf/iclr/TianKI20}             & 67.1     &50.4           \\
GMC~\cite{DBLP:conf/icml/PoklukarVYM0K22}             & 64.1          &53.1       \\ \hline
CRCD~\cite{DBLP:conf/cvpr/ZhuTCYLRYW21}       & 69.6          & 52.4          \\
CCL~\cite{DBLP:conf/cvpr/ChenXKSA21}          & 68.9          & 51.9          \\ \hline
AMID (Ours)                                   & \textbf{73.8} & \textbf{53.6} \\
    \hline
\end{tabular*}}
\label{tab:recog}
\vspace{-5pt}
\end{table}

\subsection{Comparison with State-of-the-arts}
\textbf{Video recognition.} We compare AMID with two types of SOTA cross-modal KD methods, \ie intra-sample methods including CRD~\cite{DBLP:conf/iclr/TianKI20}, GMC~\cite{DBLP:conf/icml/PoklukarVYM0K22}, and inter-sample methods including CCL~\cite{DBLP:conf/cvpr/ChenXKSA21}, CRCD~\cite{DBLP:conf/cvpr/ZhuTCYLRYW21}. The results are summarized in \cref{tab:recog}. Here "Baseline" represents the model directly trained with the visual modality only, representing a lower bound for the tasks. It can be seen that AMID significantly outperforms all the compared methods, suggesting that it is more effective and consistent in learning information from auxiliary modalities. AMID outperforms the second-best method by $4.2\%$ and $0.5\%$ on the two datasets, respectively. 

\begin{table}[t]
\setlength{\abovecaptionskip}{0cm}
\centering
\caption{$\mathrm{R@K(K}=1,5)$ of video retrieval on UCF51 and ActivityNet.}
\setlength{\tabcolsep}{1.6mm}\scalebox{1.0}{
\begin{tabular*}{0.77\linewidth}{l|cc|cc}
\hline
& \multicolumn{2}{c|}{UCF51}    & \multicolumn{2}{c}{ActivityNet} \\ \hline
Metric                                         & R1            & R5            & R1             & R5             \\ \hline
Baseline                                      & 64.0          & 71.9          & 41.5           & 63.9           \\ \hline
CRD~\cite{DBLP:conf/iclr/TianKI20}             &66.0       &72.3           &43.6         & 65.5      \\
GMC~\cite{DBLP:conf/icml/PoklukarVYM0K22}      &61.8       &71.0           &46.1            &\textbf{67.7}            \\ \hline
CRCD~\cite{DBLP:conf/cvpr/ZhuTCYLRYW21}       & 68.2        & 75.2          & 45.7           & 67.0           \\
CCL~\cite{DBLP:conf/cvpr/ChenXKSA21}          & 67.0       & 71.7          & 46.2  & 64.8           \\ \hline
AMID (Ours)     & \textbf{73.6} & \textbf{77.1} & \textbf{46.4} & 67.2  \\ 
\hline
\end{tabular*}
}
\label{tab:retri}
\end{table}

\begin{figure}[tp]
    \centering
    \includegraphics[width=0.98\linewidth,height=4.3cm]{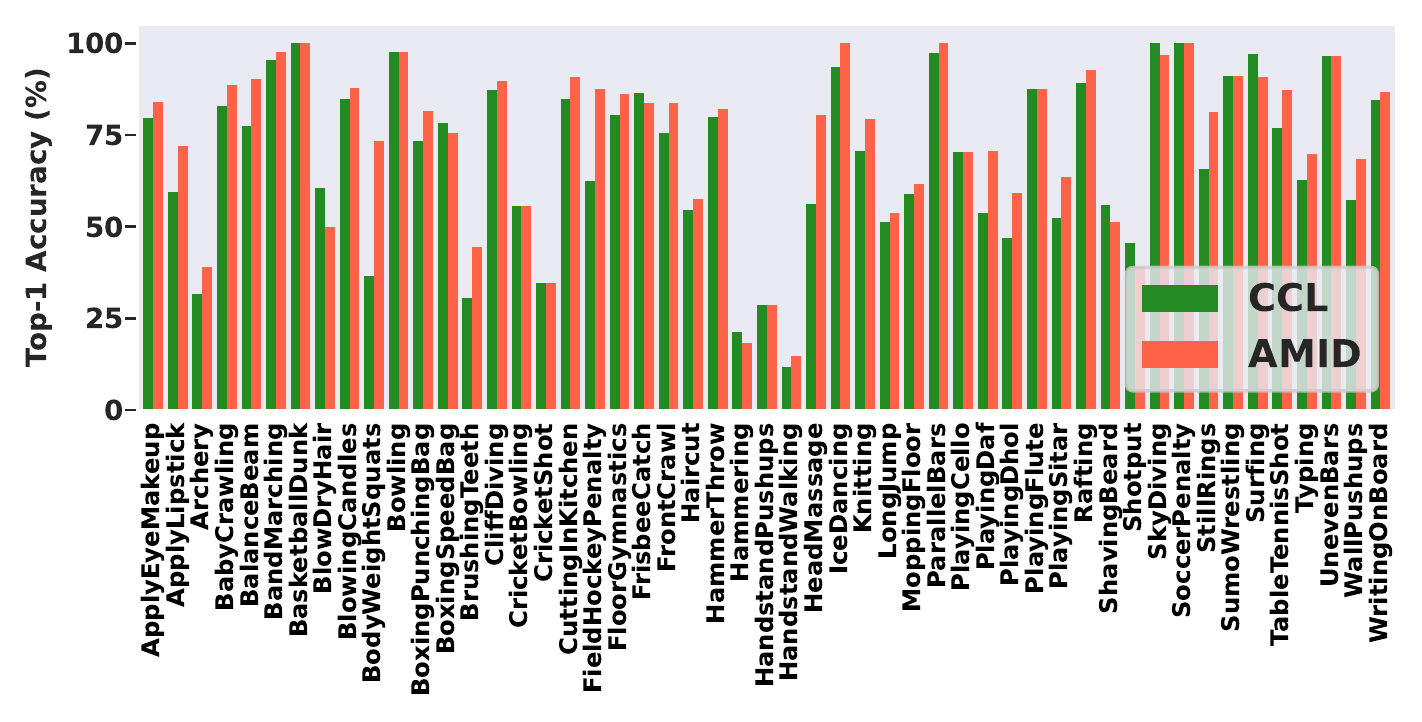}
    \caption{The per-class Top-1 accuracy (\%) of video recognition on UCF51 comparing AMID with CCL.}
    \label{fig4}
    \vspace{-5pt}
\end{figure}

To further investigate the breakdown of performance, we compare AMID with CCL on UCF51 and give the results of class-wise Top-1 accuracy in~\cref{fig4}. It can be seen that AMID achieves better performance than CCL for most of the classes, although for a few of them it is comparable to or slightly worse than CCL. The classes where AMID is of inferior performance are mainly those audio events that are not or weakly correlated with the videos, such as `FrisbeeCatch' and `SkyDiving'. In these classes, the auxiliary modal information is less helpful. Thus, AMID, which is proposed for better cross-modal knowledge distillation, is not very effective for these classes.

\textbf{Video retrieval.} \cref{tab:retri} presents the results of video retrieval on the UCF51 and ActivityNet datasets, which is a more fine-grained task to evaluate the discriminability of video representations. The results show that AMID consistently outperforms almost all competing approaches in terms of R@1 and R@5. 

\begin{table}[t]
\setlength{\abovecaptionskip}{0cm}
\caption{Performance comparison under different setups on IEMOCAP, where A-V, L-V, AL-V mean distillation from acoustic to visual, textual to visual, and acoustic and textual to visual.}
\setlength{\tabcolsep}{3.5mm}
\centering
\begin{tabular}{l|c|cc}
\hline
\multicolumn{1}{l|}{Setup}   & Method  &WA   & UA    \\ \hline
\multicolumn{1}{l|}{V} & Baseline & 51.2 & 50.8    \\ \hline
\multirow{4}{*}{A-V}        
&GMC~\cite{DBLP:conf/icml/PoklukarVYM0K22}&51.7   &50.9      \\
& CRCD~\cite{DBLP:conf/cvpr/ZhuTCYLRYW21}& 51.8 & 50.2  \\
&  CCL~\cite{DBLP:conf/cvpr/ChenXKSA21} & 51.5 & 51.4 \\
& AMID (Ours) & \textbf{52.5} & \textbf{51.7}  \\
\hline
\multirow{4}{*}{L-V} 
&GMC~\cite{DBLP:conf/icml/PoklukarVYM0K22} &50.9 & 50.5     \\
& CRCD~\cite{DBLP:conf/cvpr/ZhuTCYLRYW21} & 50.2 & 49.9  \\
& CCL~\cite{DBLP:conf/cvpr/ChenXKSA21}& 52.7 & 51.8 \\
      & AMID (Ours)   & \textbf{53.0} & \textbf{52.4}    \\   \hline

\multirow{4}{*}{AL-V} 
&GMC~\cite{DBLP:conf/icml/PoklukarVYM0K22} &51.9 & 50.5    \\
& CRCD~\cite{DBLP:conf/cvpr/ZhuTCYLRYW21}&52.0 & 49.4   \\
  & CCL ~\cite{DBLP:conf/cvpr/ChenXKSA21}& 53.2 & 51.4   \\
 & AMID (Ours)  & \textbf{53.8} & \textbf{52.7} \\ \hline
\end{tabular}
\label{tab:emo}
\end{table}

\textbf{Emotion classification.} 
\cref{tab:emo} compares AMID with three state-of-the-art methods, CCL, CRCD and GMC, for emotion classification. We conduct experiences on IEMOCAP, a three-modality dataset, to further evaluate the performance of AMID in different auxiliary modalities. It is worth noting that the "Baseline" model is trained directly with only the visual modality. The results demonstrate that AMID outperforms the Baseline model for all setups, where the UA metrics increase by 0.9\%, 1.6\% and 1.9\% respectively, while the other three methods are not always effective. Besides, AMID outperforms CCL, CRCD and GMC on almost all the setups.

\subsection{Ablation Study}
\subsubsection{Effectiveness of MI and conditional entropy}
In order to verify the effectiveness of our proposed method for capturing as much multimodal information as possible, the impact of using different training configurations on UCF51 is studied in~\cref{tab:teacher}. Here “student alone” represents the result of a model trained with the visual modality only and “teacher alone” represents that of the teacher model with the full modalities. These two cases show the initial performance of the student and the teacher without distillation, respectively. When optimizing $\mathcal{L}_{MI_s}$ alone for the distillation, although the student model can achieve performance gains, from 66.7\% to 69.8\%, it degrades the teacher's performance, from 74.3\% to 71.2\% and thereby reducing the upper bound of the student's performance. Alternatively, more informative multimodal representations can be obtained by maximizing the lower bound on $I(M;A)$ and $I(M;S)$ simultaneously ($\mathcal{L}_{MI_s}+\mathcal{L}_{MI_a}$), since a better teacher can improve the distillation performance. \cref{tab:teacher} shows that this improves the accuracy of the student from 66.7\% to 72.7\% and also improves that of the teacher from 74.3\% to 76.0\%. Besides, adding $\mathcal{L}_{adv}$ (\ie, AMID) to $\mathcal{L}_{MI_s}+\mathcal{L}_{MI_a}$ does improve the performance from 72.7\% to 73.8\%, and 
the performance gap of AMID is smaller than $\mathcal{L}_{MI_s}+\mathcal{L}_{MI_a}$, which indicates that AMID achieves better alignment between the teacher and the student due to the further constraint on the gap by minimizing $H(M|S)$ without degrading the teacher's performance obviously (the teacher's performance decreases from 76.0\% to 75.7\%).

\begin{table}[t]
    \centering
    \setlength{\abovecaptionskip}{0cm}
    \caption{Ablation study for the proposed loss functions.}
    \setlength{\tabcolsep}{2.5mm}
    \begin{tabular*}{0.98\linewidth}{l|cc}
        \hline
\multirow{2}{*}{Configuration}   & \multicolumn{2}{c}{Top-1 accuracy} \\ 
                      & Student          & Teacher         \\ \hline
student alone (w/o distillation)       & 66.7           & \textbackslash \\
teacher alone (w/o distillation) & \textbackslash & 74.3           \\ \hline
$\mathcal{L}_{MI_s}$      & 69.8           & 71.2           \\
$\mathcal{L}_{MI_s}+\mathcal{L}_{MI_a}$   & 72.7           & 76.0    \\
$\mathcal{L}_{MI_s}+\mathcal{L}_{MI_a}+\mathcal{L}_{adv}$ (AMID) & 73.8 & 75.7\\ \hline
AMID with $\mathcal{L}_{nce}$      &72.3      & 74.3         \\
        \hline
        \end{tabular*}
\label{tab:teacher}
\end{table}

\begin{figure}[t]
\centering
\setlength{\abovecaptionskip}{0.1cm}
\includegraphics[width=0.6\columnwidth]{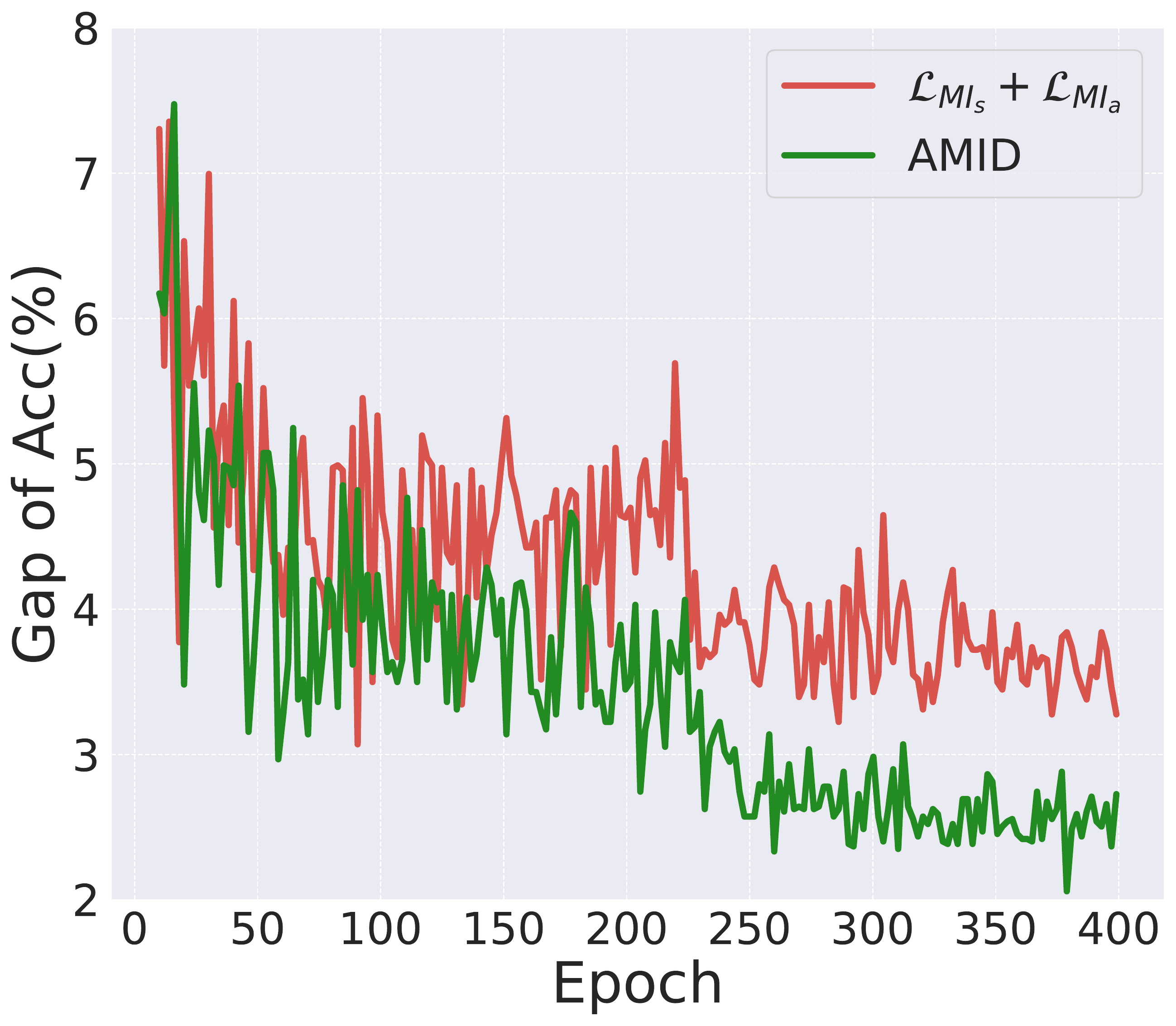} 
\caption{The accuracy gap between the teacher and student in using $\mathcal{L}_{MI_s}+\mathcal{L}_{MI_a}$ and AMID loss.}
\label{fig-delta}
\vspace{-5pt}
\end{figure}

To further understand the effectiveness of the conditional entropy minimization approach in reducing the information gap between the teacher and the student, we explore the gap of AMID and $\mathcal{L}_{MI_s}+\mathcal{L}_{MI_a}$ during training. This information gap is represented by the gap of accuracy between the teacher and student on the validation set over the training epochs. The results are shown in~\cref{fig-delta}. 
It can be seen that the accuracy gap of using AMID loss is consistently smaller than that of $\mathcal{L}_{MI_s}+\mathcal{L}_{MI_a}$, which validates the proposed adversarial learning-based approach indeed helps to narrow down the information gap.

AMID is also compared to an ablative setup using the contrastive loss $\mathcal{L}_{nce}$~\cite{DBLP:conf/cvpr/ChenXKSA21} to verify the effectiveness of the proposed loss function in maximizing the lower bound on MI. The result in~\cref{tab:teacher} shows that AMID performs much better than the baseline "AMID with $\mathcal{L}_{nce}$", improving the accuracy by 1.5\%. This validates that our proposed approach, which takes into account the correlation of samples within a class, achieves better optimization for MI.

\vspace{-7pt}
\subsubsection{Effectiveness of training modifications}
\vspace{-3pt}
To evaluate the effect of each modification applied in AMID, we conduct a set of experiments on UCF51 and give the results in~\cref{tab:ablation}. As mentioned before, using a fixed pre-trained teacher to guide the student's learning leads to an undesired gap between the teacher and the student at the beginning of training, which is unfavorable for distillation. The result affirms this, where the performance of AMID degrades from 73.8\% to 66.0\% without joint training. 
In addition, it also shows that without employing the warm-up setup described before, the recognition performance degrades by $8.0\%$. The reason may be that the target modality and multimodality representations are noisy and uncorrelated at the start of the training, which results in that the discriminators can recognize them readily and thus affecting the feedback of the informative gradient of adversarial learning.

\begin{table}[t]
    \centering
    \setlength{\abovecaptionskip}{0cm}
    \caption{Effectiveness of proposed training modifications.}
    \begin{tabular}{l|c}
    \hline
        Modification                  & Top-1 accuracy \\ \hline
        AMID                    & 73.8   \\ \hline
        with fixed teacher  & 66.0          \\
        w/o warm-up             & 65.8            \\ 
        w/o  $D_2$              & 72.8            \\
        w/o dynamic weight          & 72.7            \\ \hline
    \end{tabular}
    \label{tab:ablation}
    \vspace{-3pt}
\end{table}

We also study the effectiveness of the two discriminators of our proposed AMID by comparing AMID with two ablative cases: w/o $D_2$ and w/o dynamic weight. One constraint is removed at a time. The results of these ablation studies are 
shown in~\cref{tab:ablation}. The accuracy decreases by 1.0\% when only the fundamental discriminator $D_1$ is used, suggesting that the enhanced discriminator $D_2$ indeed helps to distill the multimodal information of samples of the same class. Moreover, it also demonstrates that using uniform weight leads to a performance drop, indicating the effectiveness of the proposed dynamic weighting scheme.
\begin{figure}[!t]
 \centering
 \includegraphics[width=0.22\textwidth]{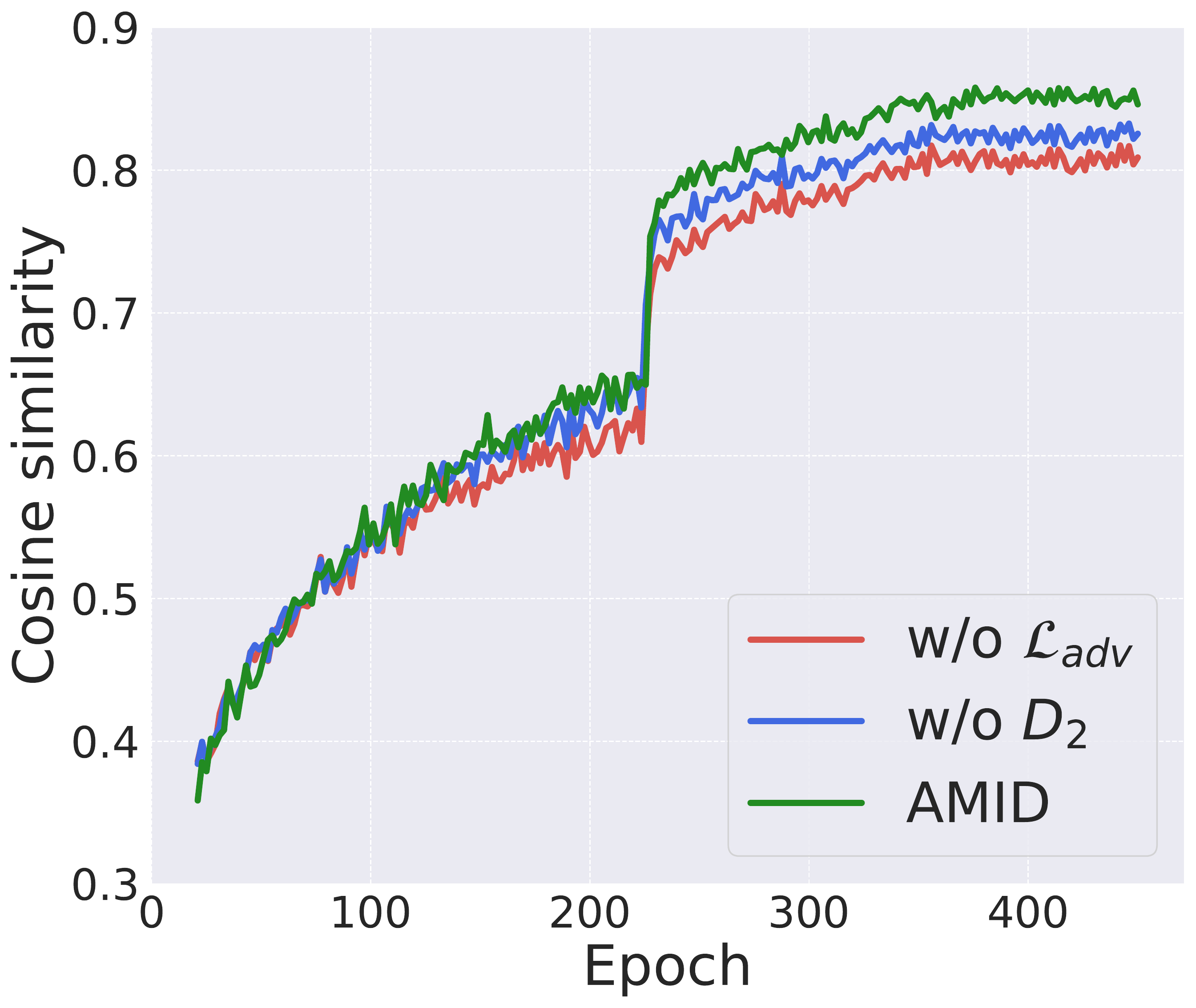}
 \includegraphics[width=0.22\textwidth]{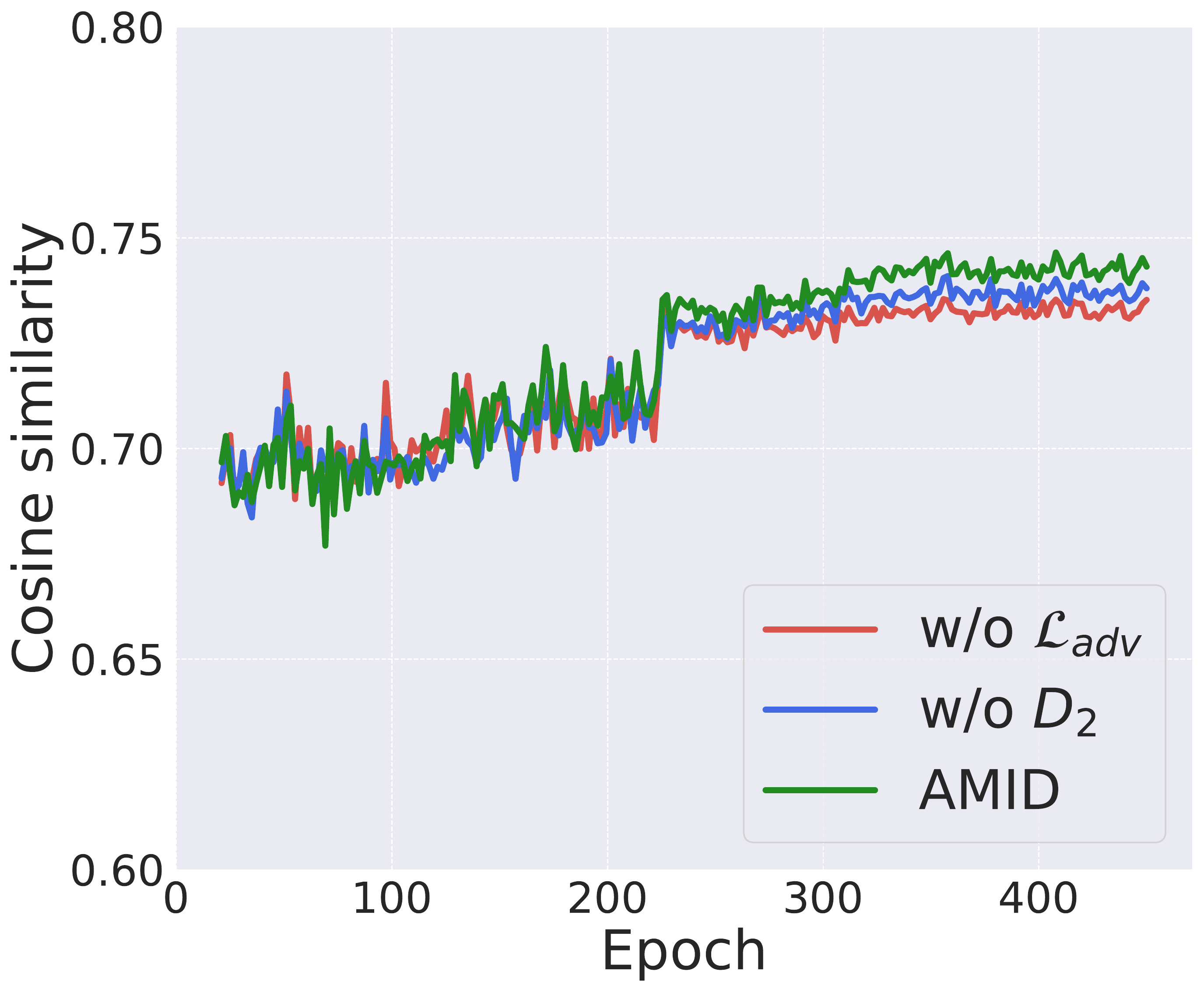} 
    \caption{(Left) The cosine similarity between the teacher's and student's representations only in the same class during training. (Right) The cosine similarity during testing.}
    \label{fig:beta_aug_k}
\vspace{-8pt}
\end{figure}

To further investigate the effectiveness of the proposed adversarial learning-based approach in transferring the inter-sample multimodal information, we compare AMID with two cases: w/o $D_2$ and w/o $\mathcal{L}_{adv}$. \cref{fig:beta_aug_k} displays the average of cosine similarity between the teacher and the student only in the same class, \ie the average of all $\Phi(M_v,S_v^{\eta_j}), j\neq0$ in \cref{eq:metric}. We can see that the cosine similarity of AMID is consistently larger than that of the other two cases and the cosine similarity of w/o $D_2$ is larger than that of w/o $\mathcal{L}_{adv}$, meaning that the proposed approach indeed helps the student to learn more compact representations for each class. 

\subsection{Hyperparameters Analysis}
\vspace{-3pt}
Sensitivity analysis studies are conducted for the hyperparameters, $\alpha_1$, $\alpha_2$ and $\alpha_3$, referenced in the loss function~\cref{eq:objective}. Similar to the previous study, we experiment on UCF51 and show the results in~\cref{tab:analysis}. The other two hyperparameters are kept the same when one changes. We test different values for $\alpha_1$ from 3.7 to 4.2, $\alpha_2$ from 1.3 to 4.8 and $\alpha_3$ from 0.7 to 1.2, all with an increment of 0.1. The results show that the hyperparameters $\alpha_1$, $\alpha_2$ and $\alpha_3$ are not sensitive within a certain range. In general, $\alpha_1$ between 3.8 to 4.1 and $\alpha_2$ between 1.4 to 1.7 works well, and we set $\alpha_3=1.0$ for all other experiments.

\begin{table}[t!]
\centering
\setlength{\abovecaptionskip}{0cm}
\caption{Top-1 accuracy (\%) on UCF51 with different hyperparameters.}
\setlength{\tabcolsep}{1.7mm}
\begin{tabular}{c|c|cccccc}
\hline
\multirow{2}{*}{$\alpha_1$} & value    & 3.7  & 3.8  & 3.9  & 4.0  & 4.1  & 4.2  \\ \cline{2-8} 
                            & accuracy & 72.6 & 73.7 & 73.7 & 73.8 & 73.8 & 73.4 \\ \hline
\multirow{2}{*}{$\alpha_2$} & value    & 1.3  & 1.4  & 1.5  & 1.6  & 1.7  & 1.8  \\ \cline{2-8} 
                            & accuracy & 71.2 & 73.7 & 73.8 & 73.8 & 73.6 & 72.4 \\ \hline
\multirow{2}{*}{$\alpha_3$} & value    & 0.7  & 0.8  & 0.9  & 1.0  & 1.1  & 1.2  \\ \cline{2-8} 
                            & accuracy & 73.3 & 73.4 & 73.7 & 73.8 & 73.4 & 73.1 \\ \hline
\end{tabular}
\label{tab:analysis}
\vspace{-6pt}
\end{table}

\section{Conclusion}
\label{sec:conc}
In this paper, we propose a generic cross-modal distillation method, AMID, to enhance the target modality representations. The proposed approach simultaneously maximizes the lower bounds on MI between the teacher and the student as well as between the teacher and an auxiliary model while minimizing the conditional entropy of the teacher given the student. For optimizing the lower bound on MI, we derive a new form and solve it by a contrastive learning-based approach. For minimizing the conditional entropy, an adversarial learning-based approach is used. Our extensive experimental evaluation demonstrates the effectiveness of AMID on three popular datasets. 

\vspace{-7pt}
\section*{Acknowledgement}
\vspace{-3pt}
This work is supported by the National Key R\&D Program of China (No. 2022ZD0160702), STCSM (No. 2251 1106101, No. 18DZ2270700, No. 21DZ1100100, No. 215 11101100), 111 plan (No. BP0719010), and State Key Laboratory of UHD Video and Audio Production and Presentation.
{\small
\bibliographystyle{ieee_fullname}
\bibliography{egbib}
}

\end{document}